\documentclass[runningheads]{llncs}
\usepackage{graphicx}
\usepackage{booktabs}
\usepackage{amsmath}
\usepackage{amssymb}
\usepackage{mathtools}
\usepackage{caption}
\usepackage{subcaption}
\usepackage{multirow}
\usepackage{ulem}
\usepackage{xcolor}
\usepackage{soul}
\usepackage[misc]{ifsym}

\begin{document}

\newpage

\title{Efficient Uncertainty Estimation in Spiking Neural Networks via MC-dropout}
\titlerunning{Efficient Uncertainty Estimation for SNNs}
%
\author{Tao Sun\inst{1} \and
Bojian Yin\inst{1} \and
Sander Boht\'e\inst{1, 2, 3}\textsuperscript{(\Letter)}}
\authorrunning{T. Sun et al.}
%
\institute{CWI, Machine Learning Group, Amsterdam, The Netherlands \\
\email{\{tao.sun,byin,sbohte\}@cwi.nl}
\and
Rijksuniversiteit Groningen, Groningen, The Netherlands\and University of Amsterdam, Amsterdam, The Netherlands}
\maketitle              

\begin{abstract}

Spiking neural networks (SNNs) have gained attention as models of sparse and event-driven communication of biological neurons, and as such have shown increasing promise for energy-efficient applications in neuromorphic hardware. As with classical artificial neural networks (ANNs), predictive uncertainties are important for decision making in high-stakes applications, such as autonomous vehicles, medical diagnosis, and high frequency trading.  Yet, discussion of uncertainty estimation in SNNs is limited, and approaches for uncertainty estimation in artificial neural networks (ANNs) are not directly applicable to SNNs. Here, we propose an efficient Monte Carlo(MC)-dropout based approach for uncertainty estimation in SNNs. Our approach exploits the time-step mechanism of SNNs to enable MC-dropout in a computationally efficient manner, without introducing significant overheads during training and inference while demonstrating high accuracy and uncertainty quality.

\end{abstract}

\section{Introduction}
Inspired by the brain's event-driven and sparse communication, spiking neural networks (SNNs) are enabling applications with high energy-efficiency in the form of neuromorphic computing \cite{schuman2022opportunities}. Analogous to biological neurons, spiking neurons in SNNs communicate using discrete spikes, and time stepping is typically used to account for the evolution of these neurons' internal state as a response to impinging and emitted spikes.  With recent advances in architectures and training methods, SNNs now achieve performance comparable to their artificial neural network (ANN) counterparts in many tasks \cite{yin2021accurate,yin2023fptt,fang2021incorporating}. 

To employ SNNs in the real-world however, accurate predictions have to be paired with high-quality uncertainty estimation to enable decision-making in high-stakes applications such as autonomous vehicles, medical diagnosis, and high frequency trading \cite{yarin2016uncertainty}: uncertain predictions in these applications may need to be reviewed by human experts for final decisions. In ANNs, predictive uncertainties in classification models are commonly represented by predictive distributions \cite{lakshminarayanan2017simple}. While evidence suggests that the brain performs a form of Bayesian inference based on uncertainty representations \cite{pouget2013probabilistic}, the literature on uncertainty in SNNs is relatively limited and primarily concentrates on the sampling of probabilistic distributions, typically from a neuroscience perspective \cite{Savin2014-qo,Jang2022-eu}.

Approaches for uncertainty estimation in classical deep learning models can be divided into two groups: deterministic methods and Bayesian methods \cite{gawlikowski2021survey}. With a deterministic method, a model learned from training data is essentially a point estimate of the model's parameters. In a deterministic deep network, each predictive distribution is estimated by a single forward propagation followed by the softmax function. Yet, although it is feasible to infer uncertainty with deterministic methods, these methods are known to be prone to output overconfident estimation \cite{lakshminarayanan2017simple,gawlikowski2021survey}. In contrast, a Bayesian network learns the posterior distribution of parameters in the network rather than depending on a single setting of parameters. The probability outputs of a Bayesian method can be analytically obtained by marginalizing the likelihood of the input with the estimated posterior distribution; this however is generally an intractable problem. To tackle this issue, many approximation methods and non-Bayesian methods have been introduced \cite{gawlikowski2021survey}. Example of these methods like Monte-Carlo-dropout (MC-dropout) \cite{gal2016dropout} and deep ensembles \cite{lakshminarayanan2017simple} achieve excellent performance in terms of uncertainty estimation quality, either by repeatedly carrying out inference for each sample in perturbed versions of the network, or by training a collection of networks and then carrying out inference in each network. 


\begin{figure}[t!]
     \centering
     \begin{subfigure}[b]{0.45\textwidth}
         \centering
         \includegraphics[width=\textwidth]{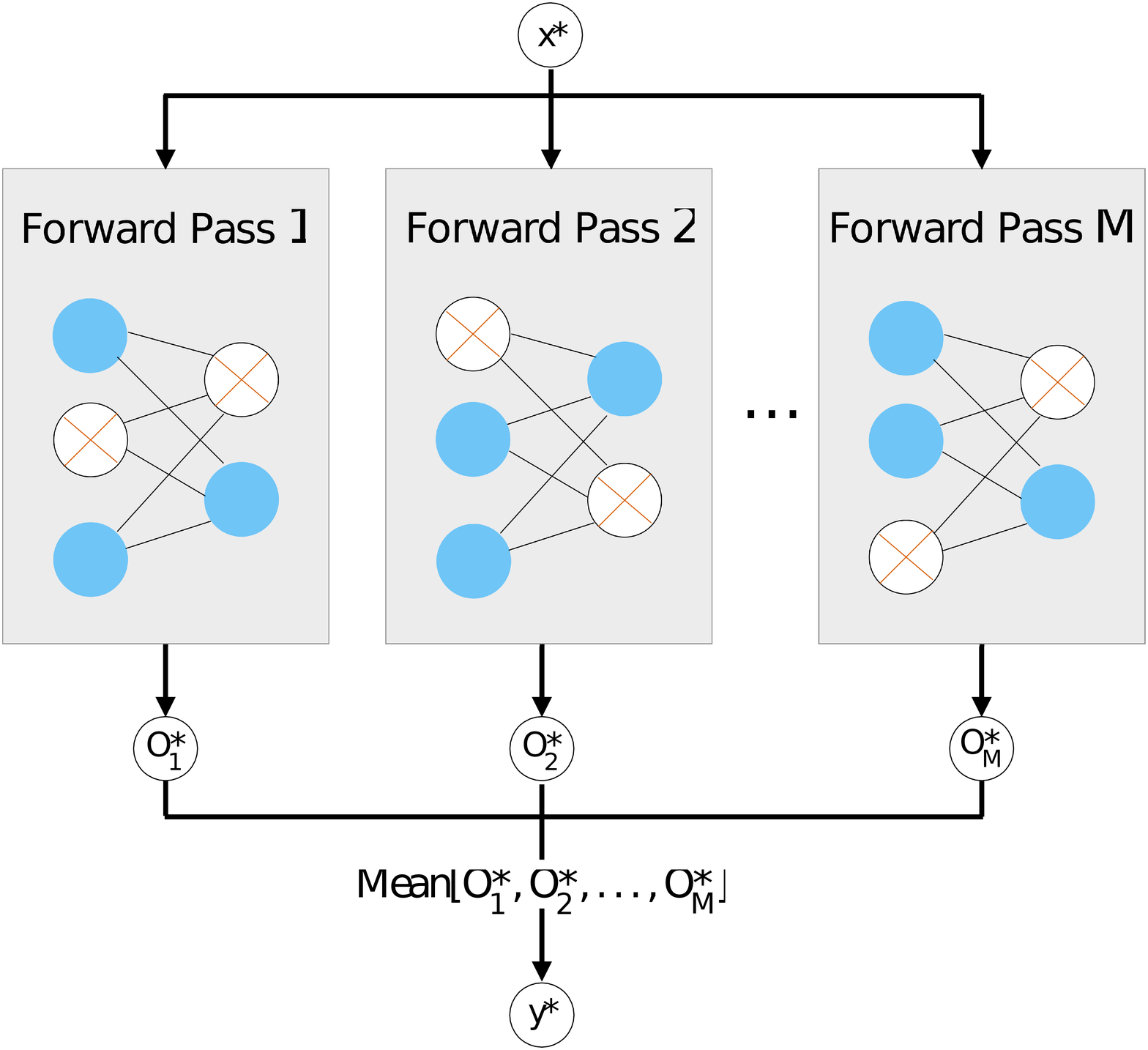}
         \caption{Inference with MC-dropout}
         \label{fig:y equals x}
     \end{subfigure}
     \hfill
     \begin{subfigure}[b]{0.45\textwidth}
         \centering
         \includegraphics[width=\textwidth]{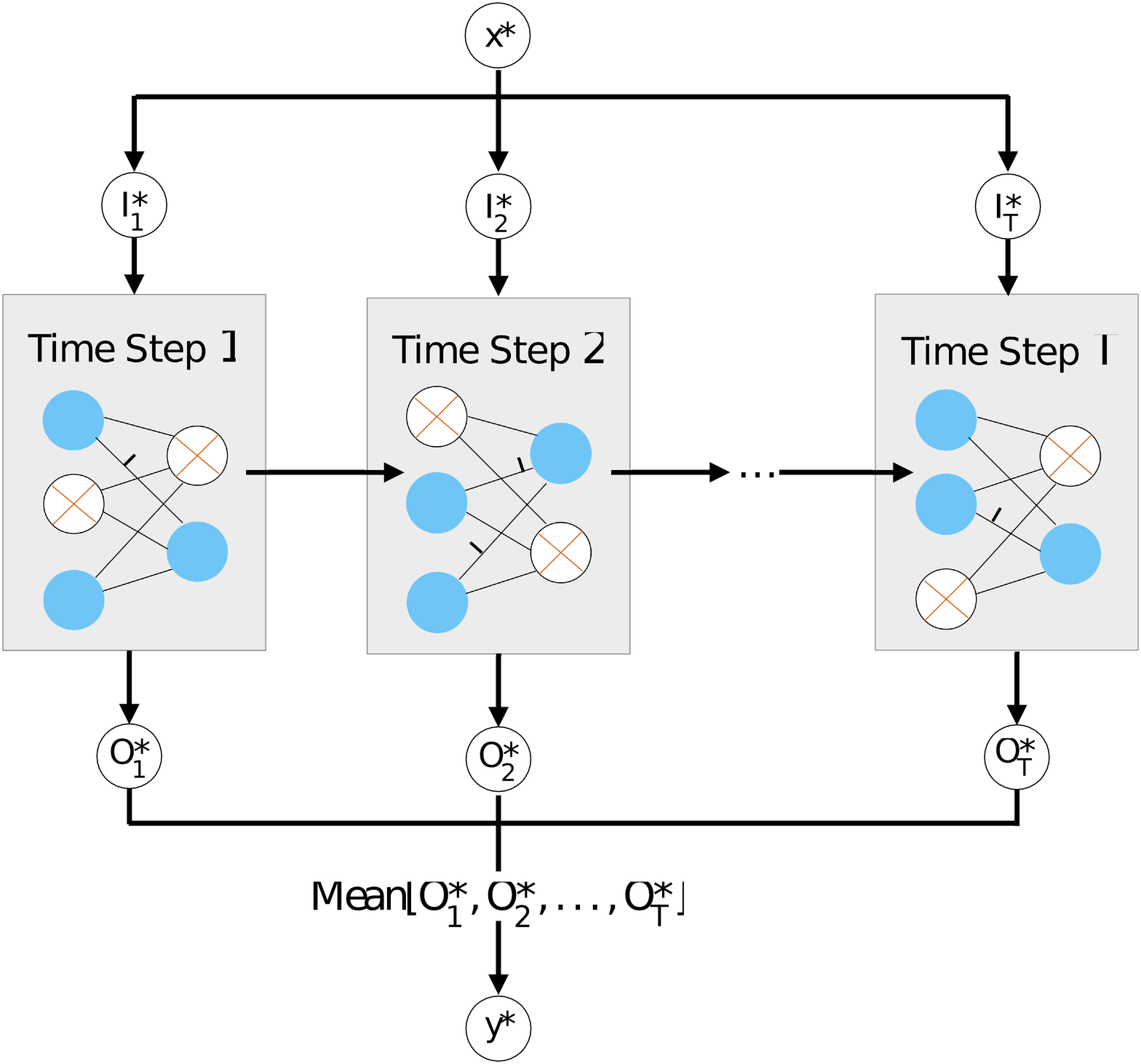}
         \caption{Inference in AOT-SNN}
         \label{fig:three sin x}
     \end{subfigure}
     \caption{(a) In ANNs, MC-dropout is performed by averaging results for a predefined number ($M$) of forward passes through a dropout-enabled network. (b) In AOT-SNNs, inference at each time step is taken as functionally equivalent to a forward pass in the MC-dropout method. As the SNN network evaluation requires $M$ time-steps already, only one effective forward pass is needed. 
     }
     \label{fig:AOT-SNN}
\end{figure}

Here, we propose an efficient uncertainty estimation approach for SNNs by exploiting their time-step mechanism. Specifically, we apply continual MC-dropout in SNNs by taking their outputs averaged over time steps as predictive distributions, where we train SNNs with a loss function that also involves their time steps: \textbf{A}verage-\textbf{O}ver-\textbf{T}ime-SNNs (AOT-SNNs, Figure \ref{fig:AOT-SNN}). In AOT-SNNs, we take inference of each time step as functionally equivalent to a forward pass in the standard MC-dropout method. Since only one forward pass is needed in inference, the computational overhead for AOT-SNNs is significantly reduced relative to the MC-dropout method while still allowing effective uncertainty estimation. We compare the performance of AOT-SNNs with more standard SNNs, as well as with SNNs using the classical MC-dropout approach and SNN ensembles, across multiple classification tasks. We demonstrate that for identical network architectures, AOT-SNNs achieve a significant gain over more standard SNNs in both accuracy and uncertainty quality while being much more computationally efficient.


\section{Background} \label{sec:bg}
\subsection{Problem Setup}
We assume a training dataset $\mathcal{D}$ that consists of $\mathcal{N}$ i.i.d data points $\mathcal{D}=\{\mathbf{X}, \mathbf{Y}\} = \{\mathbf{x}_n, y_n\}_{n=1}^N$, where $\mathbf{x}_n \in  \mathbb{R}^d$ and the true label $y_n \in \{1, \ldots, K\}$. Given a sample $\mathbf{x}_n$, a neural network outputs the probabilistic predictive distribution $p_{\omega}(y_n\vert\mathbf{x}_n)$, where $\omega$ is the parameters of the network.

A number of non-Bayesian methods achieving excellent performance in term of uncertainty estimation have been proposed, among which are deep ensembles \cite{lakshminarayanan2017simple} and post-hoc calibration methods \cite{guo2017calibration}. Deep ensembles are considered a ``gold standard'' for uncertainty estimation \cite{wilson2020bayesian}, while a set of models are trained with a proper scoring rule as the loss function. At inference time, the output of all models are then combined to obtain a predictive distribution. Post-hoc calibration methods, such as temperature scaling \cite{guo2017calibration}, involve the re-calibration of probabilities using a validation dataset and achieve excellent calibration performance in the i.i.d test dataset.  

\subsection{Bayesian Neural Networks and MC-Dropout Approximation}
In a Bayesian neural network, the predictive distribution for a sample $\mathbf{x}$ is given by: 
\begin{align} \label{eq:bayesian}
    p(y \vert \mathbf{x}, \mathcal{D}) &= \int p(y \vert \mathbf{x}, \omega)p(\omega \vert \mathcal{D})d\omega.
\end{align}
The posterior distribution, $p(\omega \vert \mathcal{D})$ or $p(\omega\vert \mathbf{X}, \mathbf{Y})$, of the parameters $\omega$ can be computed by applying Bayes' theorem
\begin{equation} \label{eq:posterior}
    p(\omega\vert \mathbf{X}, \mathbf{Y}) = \frac{p(\mathbf{Y} \vert \mathbf{X}, \omega)p(\omega)}{p(\mathbf{Y}\vert \mathbf{X})}.
\end{equation}
Due to the intractability of the normalizer in (\ref{eq:posterior}), the posterior distribution $p(\omega \vert \mathcal{D})$ and the predictive distribution $p(y \vert \mathbf{x}, \mathcal{D}))$ usually cannot be evaluated analytically. A variety of approximation methods have been introduced to tackle this issue \cite{mackay1992bayesian,graves2011practical}. One such approximation is the MC-dropout method, which is often taken as a baseline model in uncertainty estimation \cite{lakshminarayanan2017simple,ovadia2019can} due to its feasibility and relatively good performance. 

Dropout \cite{srivastava2014dropout} is a simple but effective technique used in deep learning models to prevent overfitting. In the MC-dropout method, dropout is applied before each weight layer of a neural network in both \textbf{training} and \textbf{testing}. The predictive distribution calculation with the MC-dropout method is performed by averaging results over a predefined number of forward passes through a dropout-enabled network. Gal \& Gharamani \cite{gal2016dropout} showed that neural networks with such configuration can be viewed as an approximation to a Bayesian method in the form of \textit{deep Gaussian processes} \cite{damianou2013deep}. 

Either MC-dropout models or deep ensembles involves multiple forward propagation passes in inference. As a result, when naviely applied to SNNs, the computational and energy costs becomes relatively high due to the necessity of repeatedly running SNNs for multiple times during inference.

\subsection{Source and Quality of Predictive Uncertainty} \label{sec:howtoeval}
The only source of predictive uncertainty of deterministic methods is from the noisy data. Uncertainty in a Bayesian method comes from both data and defects of the model itself \cite{gawlikowski2021survey}: uncertainty caused by data is referred to as \textit{data uncertainty}, while uncertainty caused by defects of the model itself is referred to as \textit{model uncertainty}.

The quality of predictive uncertainties can be measured from two aspects \cite{lakshminarayanan2017simple}. The first concerns uncertainty quality on in-distribution data, where test data and training data share the same distribution. The second aspect evaluates generalization of uncertainty on domain-shifted data. While certain post-hoc calibration methods may generate accurate predictive probabilities for i.i.d data, their effectiveness in predicting uncertainty for domain-shifted data is not ensured \cite{ovadia2019can}. For both aspects, model calibration is examined as the indication of uncertainty quality \cite{ovadia2019can}. For classification tasks, accuracy and calibration are two evaluation measures that are mutually orthogonal \cite{lakshminarayanan2017simple}. Accuracy, defined as the ratio of corrected classified examples to total number of examples, measures how often a model correctly classifies; calibration measures the quality of predictive probability distributions \cite{lakshminarayanan2017simple} and indicates the extent to which the probability of a predicted class label reflects the real correct likelihood. A class of metrics to measure calibration is referred to as \textit{proper scoring rules} \cite{gneiting2007strictly}, which include the Brier score (BS) and negative log-likelihood (NLL); another calibration metrics is the \textit{Expected Calibration Error} (ECE) \cite{guo2017calibration}, which is a scalar summary statistic of calibration that approximates miscalibration. Although the definition ECE is intuitive and thus widely used, it is not a perfect metric for calibration because optimal ECE values can be generated by trivial solutions \cite{ovadia2019can}; see the Appendix for details on proper scoring rules and ECE. 


\subsection{SNN}
SNNs typically work with the same types of network topologies as ANNs, but computation in SNNs is distinct. SNNs use stateful and binary-valued spiking neurons, rather than the stateless and analog valued neurons of ANNs. As a result, unlike synchronous computation in ANNs, inference in SNNs is in a iterative form through multiple time steps $t = 0, 1, ..., T$: in each time step $t$, the membrane potential of a spiking neuron $U(t)$ is affected by the impinging spikes from connecting neurons emitted at time step $t-1$, and the past potential $U(t-1)$. Once the membrane potential $U(t)$ reaches a threshold $\theta$, the neuron itself emits a spike. Such sparse and asynchronous communications between connected neurons is key to enabling SNNs to achieve high energy-efficiency.

\subsubsection{LIF Neurons}
Various spiking neuron models exist, ranging in complexity from the detailed Hodgkin-Huxley model to the simplified Leaky-Integrated-and-Fire (LIF) neuron model \cite{Gerstner2002-wd}. The latter is widely used in SNNs, as it is interpretable and computationally efficient. Resembling an RC circuit, the LIF neural model is represented as:
\begin{equation} \label{eq:lif_diff}
\tau\frac{dU}{dt} = -U + RI.
\end{equation}
where $I$ and $R$ are the current and input resistance, and $\tau$ is the time constant of the circuit. The discrete approximation of (\ref{eq:lif_diff}) can be written as:
\begin{equation}
    u^t_i = \lambda u^{t-1}_i + \sum_{j} w_{ij}s^{t}_j - s_i^{t-1}\theta,
\end{equation}
\begin{equation}
  s^t_i = \left\{
  \begin{array}{lr}
    1, & \text{if } u^t_i > \theta\\
    0, &  \text{otherwise}
  \end{array}
\right.
\end{equation}
where $u_i$ is the membrane potential of a neuron $i$, $\lambda$ denotes the leaky constant ($<1$) for the membrane potential, $w_{ij}$ represents the weight connecting the neuron $i$ and its pre-synaptic neuron $j$, and $s_i$ indicates whether a neuron spikes. 

With the introduction of surrogate gradient methods \cite{neftci2019surrogate,yin2021accurate} and learnable LIF neurons \cite{yin2021accurate,fang2021incorporating}, both trainability and performance of SNNs have been improved dramatically. 

\section{Methods}


Here, we present our proposed AOT-SNNs. We first explain how we efficiently apply MC-dropout to SNNs, and then introduce the loss function used in AOT-SNNs, which is based on the mean output values over time steps. Lastly, we explain the network architecture we use to demonstrate AOT-SNNs in practice.

\subsection{Efficient MC-dropout in SNNs}
As noted, the standard MC-dropout method runs a test sample multiple times in a model with dropout enabled, and takes the output of these forward passes as the final predictive distribution. Thus applied, the MC-dropout method achieves satisfactory performance in predictive uncertainty in ANNs. 

In principle, thus defined MC-dropout can be applied directly to SNNs, as {\it MC-dropout SNN}. This, however, can result in inefficient inference, mainly due to the time-step mechanism in SNNs. As each neuron's activation in an SNN is a continuous process over time, an SNN typically has to be run for multiple time steps to perform inference. Naively performing inference of a single sample in an SNN with MC-dropout would mean running multiple forward passes of a sample through a network where each individual pass entails the evaluation of multiple time steps. This will obviously be computationally expensive. 

As an alternative, we propose to leverage SNN time-step mechanism by enabling MC-dropout in AOT-SNNs during a single evaluation. Specifically, we compute predictive distributions in a dropout-enabled AOT-SNN by averaging outputs of its multiple time steps. In this view, each time step in an AOT-SNN is weakly equivalent to a forward pass in the standard MC-dropout method. This approach requires only one forward pass during inference and thus significantly lowers computational costs. 

\subsection{Loss Function}
Loss functions in many current high-performing SNN learning algorithms \cite{yin2021accurate,fang2021incorporating,Rathi2020Enabling,yue2023hybrid} are computed based on the output values of last time step, and we will refer such loss functions as \textit{last-time-step} loss, resulting in Last-Time-Step-SNNs ({\it LTS-SNNs}). The last-time-step loss is written as:
\begin{equation}
    \qquad L = l(T)
\end{equation}
where $l(t)$ is the loss function computed on the output values of the time step $t$. 
Since the last-time-step loss is not compatible with the proposed uncertainty estimation approach in AOT-SNNs, we introduce the \textit{average-over-time} loss, which calculates its output by averaging over multiple time steps:
\begin{equation}
L = \frac{1}{T}\sum_{t=1}^T l(t).
\end{equation}
By combining the average-over-time loss with MC-dropout, we expect that the quality of uncertainty estimation for our approach will be improved, particularly in comparison to LTS-SNNs trained by the last-time-step loss, as AOT pushes SNNs to correctly classify as much as possible at every time step, while LTS-SNNs do not.

For $l(t)$, either negative log-likelihood (NLL) loss or the mean squared error (MSE) loss \cite{fang2021incorporating} can be used. Here, we use the MSE loss, as we find that in practise the NLL loss causes practice a disconnect between NLL and accuracy, which is an indication of miscalibration \cite{guo2017calibration}.  

\subsection{Network Architecture}
We use AOT-SNNs with a network architecture very similar to the high-performing PLIF networks in  \cite{fang2021incorporating}. These networks are composed of a \textit{spiking encoder network} and a \textit{classifier network}. The spiking encoder network consists of multiple downsampling modules. Each downsampling module has a certain number of convolution blocks and a pooling layer ($kernel \, size=2, stride=2$). The convolution block is composed of a convolution layer ($kernel \, size=3, stride=1, padding=1$), a batch normalization layer, and a spiking neuron layer. 

Our classifier network is slightly modified from \cite{fang2021incorporating} and includes a fully-connected layer, a spiking neuron layer, another fully-connected layer, which is then followed by a readout integrator layer. Unlike the original PLIF networks that classify using relatively coarse summed rate-coding collected from a population of output neurons, probabilities of AOT-SNNs are computed based on the membrane potentials of readout integrator neurons as in \cite{yin2021accurate}. This modification enable AOT-SNNs to achieve better uncertainty estimation performance compared to corresponding standard PLIF networks while obtaining similar accuracy. In the spiking neuron layers, PLIF neurons \cite{fang2021incorporating} are used, where the time constants $\tau$ are learned and shared by neurons within the same layer. Note that dropout is applied to the neurons' output spikes, and input data is directly injected into the network as current into the input neurons.   


\section{Experiments}

We performed a series of experiments to compare AOT-SNNs to LTS-SNNs, as well as MC-dropout SNNs and also with the `gold standard' of SNN ensembles, across multiple classification tasks. As a proof of concept, we first applied this approach to the MNIST dataset. Second, we experiment on the CIFAR-10 dataset to compare our models with corresponding LTS-SNNs. Additionally, we reported and analyzed results on the CIFAR-100 dataset. Furthermore, we carried out an ablation study where we characterized the uncertainty properties of AOT-SNNs with regard to dropout rates and dropout types.

\subsection{Experimental Setup}

In our experiments, LTS-SNNs used the same layer structure as their corresponding AOT-SNNs. However, they differ in that LTS-SNNs used the predictive distribution output by the last time step and were trained with the last-time-step loss. Note that dropout is not enabled during inference in LTS-SNNs. Enabling dropout would lead to notably weak performance for LTS-SNNs, similar to that of ANNs. All the MC-dropout SNNs and SNN ensembles are based on their corresponding LTS-SNNs.

The Adam optimizer was used, with a cosine annealing learning rate scheduler, whose initial learning rate is 0.001 and $T_{max}$ is 64. The default dropout rate used is 0.5. For the MINIST dataset, we used a batch size of 150, while the batch sizes were 60 for CIFAR-10 and 15 for CIFAR-100. The number of epochs used for each dataset were 200 (MNIST), 300 (CIFAR-10), and 300 (CIFAR-100).

\subsection{MNIST}

\begin{table}[!t]
\caption{Performance comparisons between the AOT-SNN and its corresponding LTS-SNN on the MNIST dataset (mean$\pm$std across 5 trials). The numbers after the model names represent time steps.}
\label{tab:mnist}
\begin{center}
\begin{small}
\begin{sc}
\begin{tabular}{ccccc}
\toprule
Model & Accuracy (\%) $\uparrow$ & BS $\downarrow$ & NLL $\downarrow$ & ECE $\downarrow$ \\
\midrule
AOT-SNN (8) & 99.54$\pm$0.030 & 7.0e-4$\pm$4.3e-5 & 0.0144$\pm$7.6e-4 & 0.0012$\pm$3.4e-4   \\
\midrule
LTS-SNN (8) & \text{99.37$\pm$0.080} & \text{1.0e-3$\pm$1.0e-4} & \text{0.021$\pm$0.0025} & \text{0.004$\pm$0.0011} \\
\bottomrule
\end{tabular}
\end{sc}
\end{small}
\end{center}
\end{table}

\begin{figure}[!t]
\includegraphics[width=\textwidth]{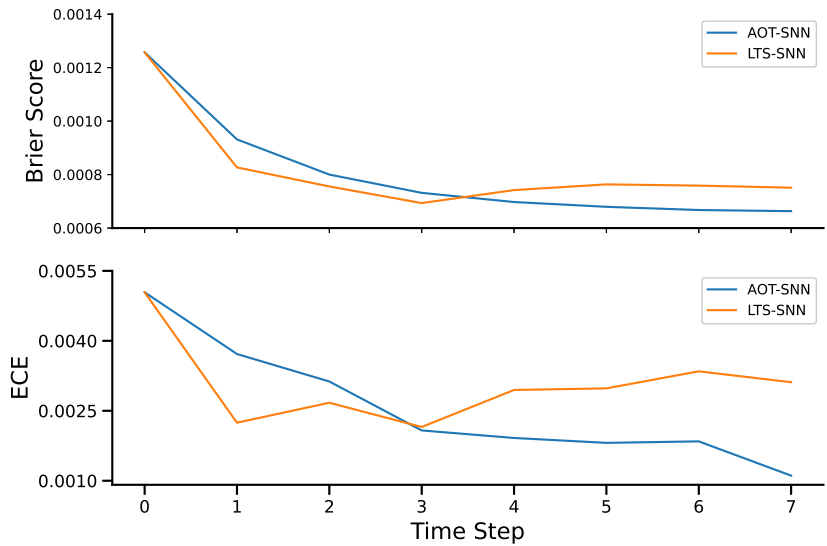}
\caption{Uncertainty performance for each time step of the AOT-SNN model and its corresponding LTS-SNN on the MNIST dataset. The results of metrics are averaged over the entire test dataset. 
}
\label{fig:mnist_avg}
\end{figure}

The spiking encoder network for the MNIST dataset has two downsampling modules, each of which includes only one convolution block. In Table \ref{tab:mnist}, we compared the AOT-SNN and its corresponding LTS-SNN, both using best performing models that have eight time steps to evaluate samples. The results demonstrate that the AOT-SNN outperforms the LTS-SNN in both accuracy and the predictive uncertainty metrics, including Brier score, NLL, and ECE. For each time step of the two models, we illustrated the performance of Brier score and ECE, averaged over the entire test dataset (figure \ref{fig:mnist_avg}). Note that the mean values of these two metrics over the previous time steps were used for each time step of the AOT-SNN. The graph illustrates that during the initial three time intervals, the AOT-SNN's performance lags behind the LTS-SNN. However, performance of the AOT-SNN demonstrates a significant improvement over the LTS-SNN in both two uncertainty metrics from the fourth time step to the final one. This improvement could be attributed to the updates of AOT-SNNs compared to LTS-SNNs, where we average the outputs of dropout-enabled time steps and take them as the final output.

\subsection{CIFAR-10 and CIFAR-100}
The architectures of AOT-SNNs for the CIFAR-10 and CIFAR-100 dataset are similar. They apply the same spiking encoder network, which has two downsampling modules, each with three convolution blocks. Their classifier networks differ only in the last fully-connected layer due to their different number of ground truth classes.

\begin{table*}[!b]
\caption{Comparison on the CIFAR-10 dataset between AOT-SNNs, LTS-SNNs, MC-dropout models, and deep ensembles (mean$\pm$ std across 5 trials). The digits enclosed in brackets following the model names indicate the number of SNN time steps and the number of forward passes or models used in inference.}
\label{tab:cifar10baseline}
\begin{center}
\begin{small}
\begin{sc}
\begin{tabular}{ccccc}
\toprule
Model & Accuracy (\%) $\uparrow$ & BS $\downarrow$ & NLL $\downarrow$ & ECE $\downarrow$ \\
AOT-SNN (4, 1) & 90.2$\pm$0.26 & 0.0153$\pm$0.00030 & 0.38$\pm$0.012 & 0.040$\pm$0.0031  \\
AOT-SNN (8, 1) & 90.8$\pm$0.23 & 0.0144$\pm$0.00040 & 0.37$\pm$0.022 & 0.043$\pm$0.0041 \\
LTS-SNN (4, 1) & 88.9$\pm$0.71 & 0.017$\pm$0.0011 & 0.43$\pm$0.028 & 0.058$\pm$0.0044 \\
LTS-SNN (8, 1) &   88.5$\pm$0.60 & 0.0181$\pm$0.00081 & 0.47$\pm$0.013 & 0.067$\pm$0.0034 \\
MC-dropout SNN (4, 5) & 90.53$\pm$0.37 & 0.0140$\pm$0.00041 & 0.32$\pm$0.001 & 0.026$\pm$0.0030 \\
MC-dropout SNN (8, 5) & 90.43$\pm$0.37 & 0.0145$\pm$0.00053 & 0.35$\pm$0.013 & 0.037 $\pm$0.0014 \\
SNN Ensembles (4, 5) & 90.9 &  0.0134 & 0.2919 & 0.012 \\
SNN Ensembles (8, 5) & 90.8 &  0.0135 & 0.2967 & 0.016 \\
\bottomrule
\end{tabular}
\end{sc}
\end{small}
\end{center}
\end{table*}

\begin{table*}[!t]
\caption{In-distribution performance comparisons between AOT-SNNs and LTS-SNNs on CIFAR10 (mean$\pm$std across 5 trials).}
\label{tab:cifar10steps}
\begin{center}
\begin{small}
\begin{sc}
\begin{tabular}{cccccc}
\toprule
Model & Time steps & Accuracy (\%) $\uparrow$ & BS $\downarrow$ & NLL $\downarrow$ & ECE $\downarrow$ \\
\midrule
AOT-SNN & 2 & 89.4$\pm$0.18 & 0.0168$\pm$0.00014 & 0.417$\pm$0.0061 & 0.047$\pm$0.0023   \\
AOT-SNN& 3 & 89.7$\pm$0.26 & 0.0160$\pm$0.00027 & 0.40$\pm$0.021 & 0.044$\pm$0.0042  \\
AOT-SNN& 4 & 90.2$\pm$0.26 & 0.0153$\pm$0.00030 & 0.38$\pm$0.012 & 0.040$\pm$0.0031  \\
AOT-SNN& 5 & 90.4$\pm$0.07 & 0.0150$\pm$0.00024 & 0.39$\pm$0.026 & 0.043$\pm$0.0036  \\
AOT-SNN& 6 & 90.5$\pm$0.16 & 0.0149$\pm$0.00028 & 0.38$\pm$0.017 & 0.043$\pm$0.0030 \\
AOT-SNN& 7 & 90.2$\pm$0.34 & 0.0151$\pm$0.00043 & 0.37$\pm$0.012 & 0.043$\pm$0.0019 \\
AOT-SNN& 8 & 90.8$\pm$0.23 & 0.0144$\pm$0.00040 & 0.37$\pm$0.022 & 0.043$\pm$0.0041 \\
AOT-SNN& 9 & 90.5$\pm$0.55 & 0.0147$\pm$0.00073 & 0.37$\pm$0.024 & 0.044$\pm$0.0041 \\
AOT-SNN& 10 & 90.7$\pm$0.41 & 0.0146$\pm$0.00062 & 0.37$\pm$0.024 & 0.044$\pm$0.0052\\
\midrule
LTS-SNN & 1 & 88.2$\pm$0.47 & 0.017$\pm$0.00068 & 0.36$\pm$0.013 & 0.0138$\pm$0.0034 \\
LTS-SNN & 2 & 88.6$\pm$0.40 & 0.0180$\pm$0.00031 & 0.46$\pm$0.0085 & 0.067$\pm$0.0055 \\
LTS-SNN & 3 & 88.0$\pm$0.56 & 0.0184$\pm$0.00076 & 0.44$\pm$0.023 & 0.060$\pm$0.0030 \\
LTS-SNN & 4 & 88.9$\pm$0.71 & 0.017$\pm$0.0011 & 0.43$\pm$0.028 & 0.058$\pm$0.0044 \\
LTS-SNN & 5 & 88.4$\pm$0.27 & 0.0181$\pm$0.00047 & 0.46$\pm$0.016 & 0.063$\pm$0.0031  \\
LTS-SNN & 7 & 88.3$\pm$1.12 & 0.018$\pm$0.0014 & 0.48$\pm$0.026 & 0.068$\pm$0.0062   \\
LTS-SNN & 8 & 88.5$\pm$0.60 & 0.0181$\pm$0.00081 & 0.47$\pm$0.013 & 0.067$\pm$0.0034 \\
LTS-SNN & 9 & 88.0$\pm$0.52 & 0.0189$\pm$0.00082 & 0.49$\pm$0.025 & 0.069$\pm$0.0036 \\
LTS-SNN & 10  & 88.0$\pm$0.91 & 0.019$\pm$0.0015 & 0.49$\pm$0.046 & 0.069$\pm$0.0062 \\
\bottomrule
\end{tabular}
\end{sc}
\end{small}
\end{center}
\end{table*}

\subsubsection{CIFAR-10 held-out test dataset.}
Table \ref{tab:cifar10baseline} presents a comparison of AOT-SNNs to LTS-SNNs, MC-dropout SNNs, and SNN ensembles. While each MC-dropout SNN ran five forward passes, each SNN ensemble consisted of five models. We show results for 4 and 8 time steps, corresponding to respective best performing duration (see also Table \ref{tab:cifar10steps}). AOT-SNNs exhibit superior performance compared to LTS-SNNs, and achieve comparable accuracy to SNN ensembles while yielding slightly lower results on BS and NLL, only underperforming on ECE. In comparison to the the MC-dropout SNNs, AOT-SNNs do deliver superior accuracy and performed almost as well as BS and NLL, with only a slight loss in ECE. 

Table \ref{tab:cifar10steps} presents the results of AOT-SNNs and LTS-SNNs with time steps smaller or equal to 10. With each model trained five times, the table lists the mean and standard deviation for all the metrics. In this exhaustive comparison, we see that that AOT-SNNs significantly outperform LTS-SNNs, with all models with more than 3 time steps achieving significantly better accuracy and Brier score, with best results for 8 time steps. Moreover, almost all AOT-SNNs achieve better NLL and ECE, except for the model with a single time step (which however has considerably lower accuracy). This is in line with the finding for MNIST in the Figure \ref{fig:mnist_avg}.



\subsubsection{CIFAR-100.}

Comparing the AOT-SNN with time step eight with its corresponding LTS-SNN for CIFAR-100 (Table \ref{tab:cifar100}), we similarly find that AOT-SNNs achieve significantly better results than the LTS-SNN, in both accuracy and predictive uncertainty quality.

\begin{table*}[!t]
\caption{Performance comparisons between the AOT-SNN and the corresponding LTS-SNN on the CIFAR-100 dataset.}
\label{tab:cifar100}
\begin{center}
\begin{small}
\begin{sc}
\begin{tabular}{cccccc}
\toprule
Model & Time steps & Accuracy (\%) $\uparrow$ & BS $\downarrow$ & NLL $\downarrow$ & ECE $\downarrow$ \\
\midrule
AOT-SNN& 8 &  65.15 & 0.005028 & 1.6749 & 0.1352\\
LTS-SNN & 8 & 62.32 & 0.005333 & 1.7325 & 0.1665  \\
\bottomrule
\end{tabular}
\end{sc}
\end{small}
\end{center}
\end{table*}


\subsubsection{CIFAR-10-C: domain-shifted test dataset.}
\begin{figure}[t!]
\includegraphics[width=\textwidth]{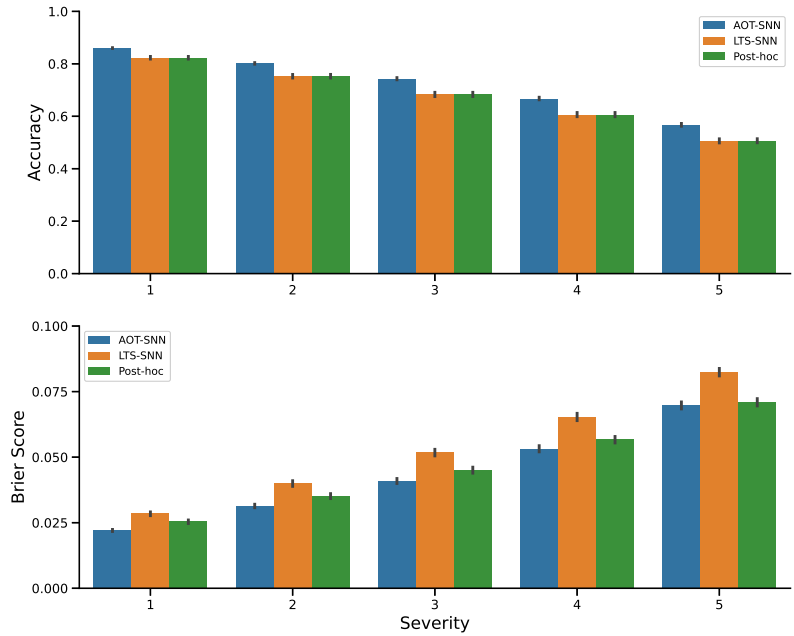}
\caption{Comparisons of the AOT-SNN model, its corresponding LTS-SNN, and the post-hoc calibrated version of the LTS-SNN on each severity level of CIFAR-10-C.} 
\label{fig:ood_robust}
\end{figure}

As mentioned earlier, the quality of predictive uncertainties need to be measured on both in-distribution held-out data and domain-shifted data. We evaluated AOT-SNNs on the CIFAR-10-C dataset \cite{hendrycks2018benchmarking}, a domain-shifted test dataset of CIFAR-10. The CIFAR-10-C dataset is designed to evaluate the robustness of image classification models against common corruptions. It contains 19 corruption types that are created by applying a combination of 5 severity levels to the original CIFAR-10 test set. The CIFAR-10-C dataset is commonly used as a benchmark to evaluate the uncertainty estimation in domain-shifted settings \cite{ovadia2019can}. We compared the performance of the AOT-SNN with eight time steps, its corresponding LTS-SNN, and the temperature scaling method that re-calibrates probabilities output by the LTS-SNN on all the severity levels of CIFAR-10-C (Figure \ref{fig:ood_robust}). The results show that the AOT-SNN outperform the LTS-SNN in all severity levels. Furthermore, in severity levels one to four, the AOT-SNN achieve better performance than the post-hoc calibrated results, while in level five, the AOT-SNN has comparable performance. Together, this shows that AOT-SNNs improve uncertainty estimation over both LTS-SNNs and the temperature scaling method also in domain-shifted settings. 

\subsubsection*{Ablation study.}

We further considered the impact of dropout rates and dropout types on the quality of uncertainty estimates of AOT-SNNs.

\paragraph{Dropout type.}

 We replaced the dropout in the LTS-SNN and our best-performing model, both of which have eight time steps, with DropConnect \cite{wan2013regularization}. Instead of  dropping the spikes like the regular dropout, DropConnect randomly drops the weights in each layer before the PLIF neuron layer. As shown in Table \ref{tab:dropconnect}, despite the slightly better performance of the LTS-SNN-DC compared to the corresponding dropout-based models (LTS-SNN), the AOT-SNN-DC outperform LTS-SNN-DC in terms of both accuracy and uncertainty quality (both models in the table have a dropout rate of 0.5). The observation suggests that DropConnect may fulfill the same function as regular dropout in AOT-SNNs.

\begin{table*}[!h]
\caption{Performance comparisons between the AOT-SNN with DropConnect and its corresponding LTS-SNN on the CIFAR-10 dataset. The numbers after the model names represent time steps.}
\label{tab:dropconnect}
\begin{center}
\begin{small}
\begin{sc}
\begin{tabular}{ccccc}
\toprule
Model & Accuracy (\%) $\uparrow$ & BS $\downarrow$ & NLL $\downarrow$ & ECE $\downarrow$ \\
\midrule
AOT-SNN (8)    & 90.8$\pm$0.23 & 0.0144$\pm$0.00040 & 0.37$\pm$0.022 & 0.043$\pm$0.0041 \\
AOT-SNN-DC (8) & 90.5$\pm$0.37 & 0.0140$\pm$0.00041 & 0.32$\pm$0.010 & 0.026$\pm$0.0030 \\
LTS-SNN (8)    & 88.5$\pm$0.60 & 0.0181$\pm$0.00081 & 0.47$\pm$0.013 & 0.067$\pm$0.0034 \\
LTS-SNN-DC (8) & 90.2$\pm$0.25 & 0.0161$\pm$0.00036 & 0.47$\pm$0.035 & 0.065$\pm$0.0041 \\
\bottomrule
\end{tabular}
\end{sc}
\end{small}
\end{center}
\end{table*}

\begin{figure}[!h]
\includegraphics[width=\textwidth]{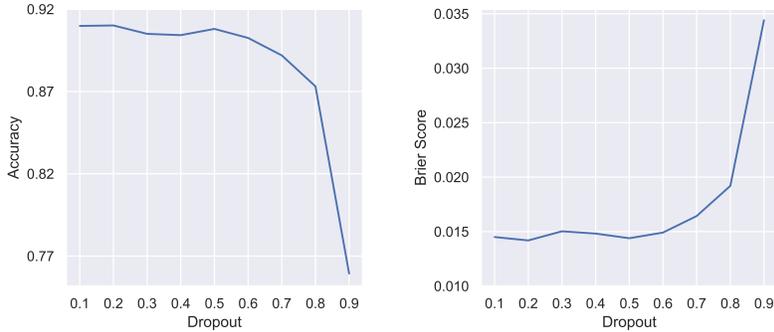}
\caption{The impact of dropout rate on performance of AOT-SNNs on the CIFAR-10 dataset. Dropout rates are ranging from 0.1 to 0.9 in increments of 0.1. 
} 
\label{fig:dropout}
\end{figure}

\paragraph{Dropout rate.}
To investigate the impact of dropout rate on performance, we tested AOT-SNNs with dropout rates ranging from 0.1 to 0.9 in increments of 0.1. These experiments were based on our best-performing model of eight time steps and trained on the CIFAR-10 dataset separately for each amount of dropout. The accuracy and Brier score were plotted in Figure \ref{fig:dropout}. The trends in accuracy, Brier score are consistent, with models having dropout rates lower than 0.5 producing flat results, followed by a decline in performance.

\section{Conclusion}
We proposed a novel and efficient approach for uncertainty estimation in spiking neural networks SNNs based on the MC-dropout method combined with an appropriate choice of loss-function. Our approach exploits the time-step mechanism of SNNs to enable MC-dropout in a computationally efficient manner, without introducing significant overheads during training and inference. We demonstrated that our proposed approach can be computationally efficient and performant in uncertainty quality at the same time. Future work could investigate the potential of our approach in more applications, such as speech processing and medical imaging.


\vspace{-0.5cm}
\subsubsection*{Acknowledgments}
TS is supported by NWO-NWA grant NWA.1292.19.298. SB is supported by the European Union (grant agreement 7202070 ``HBP'').

%
%
%

\appendix \label{appdix}
\section*{Appendix}
\subsubsection*{Proper Scoring Rules.}
A \textit{scoring rule} $S(\mathbf{p},y)$ assigns a value for a predictive distribution $\mathbf{p}$ and one of the labels $y$. A \textit{scoring function} $s(\mathbf{p},\mathbf{q})$ is defined as the expected score of $S(\mathbf{p},y)$ under the distribution $q$ 
\begin{equation}
    s(\mathbf{p},\mathbf{q}) = \sum_{y=1}^{K} q_yS(\mathbf{p}, y).
\end{equation}
If a scoring rule satisfies $s(\mathbf{p},\mathbf{q}) <= s(\mathbf{q},\mathbf{q})$, it is called a \textit{proper scoring rule}. If $s(\mathbf{p},\mathbf{q}) = s(\mathbf{q},\mathbf{q})$ implies $\mathbf{q}=\mathbf{p}$, this scoring rule is a \textit{strictly proper scoring rule}. When evaluating quality of probabilities, an optimal score output by a proper scoring rule indicates a perfect prediction \cite{ovadia2019can}. In contrast, trivial solutions could generate optimal values for an improper scoring rule \cite{ovadia2019can,gneiting2007strictly}. 

The two most commonly used proper scoring rules are Brier score \cite{brier1950verification} and NLL. Brier score is the squared $L_2$ norm of the difference between $\mathbf{p}$ and one-hot encoding of the true label $y$. NLL is defined as $ S(\mathbf{p}, y) = -\mathrm{log} p(y\vert\mathbf{x})$ with $y$ being the true label of the sample $\mathbf{x}$. Among these two rules, the Brier score is more recommendable because NLL can unacceptably over-emphasize small differences between small probabilities \cite{ovadia2019can}. Note that proper scoring rules are often used as loss functions to train neural networks. \cite{lakshminarayanan2017simple,gneiting2007strictly}.

\subsubsection*{ECE.}
The ECE is a scalar summary statistic of calibration that approximates miscalibration \cite{naeini2015obtaining,guo2017calibration}. To calculate ECE, the predicted probabilities, \\$\hat{y}_n = \mathrm{argmax}_y \mathbf{p}(y\vert\mathbf{x_n})$, of test instances are grouped into $M$ equal-interval bins. The ECE is defined as
\begin{equation}
    ECE = \sum_{m=1}^M f_m \vert o_m - e_m\vert ,
\end{equation}
where $o_m$ is the fraction of corrected classified instances in the $m^{th}$ bin, $e_m$ the average of all the predicted probabilities in the $m^{th}$ bin, and $f_m$ the fraction of all the test instances falling into the $m^{th}$ bin. The ECE is not a proper scoring rule and thus optimum ECEs could come from trivial solutions.

\vspace{-0.2cm}

\bibliographystyle{splncs04}
\bibliography{uncertainty_main}

\end{document}